\documentclass[letterpaper, 10 pt, journal, twoside]{ieeetran}

\IEEEoverridecommandlockouts                              

\usepackage{amsmath,amsfonts}
\usepackage{algorithm}
\usepackage{array}
\usepackage[
    caption=false,
    font=footnotesize,
    labelfont=sf,
    textfont=it
]{subfig}
\usepackage{textcomp}
\usepackage{stfloats}
\usepackage{url}
\usepackage{verbatim}
\usepackage{cite}

\usepackage{graphicx}
\usepackage{xcolor}

\usepackage{silence}

\usepackage{amssymb}
\usepackage{pifont}
\usepackage{multirow}
\usepackage{booktabs}
\usepackage{tabularx}
\usepackage{comment}

\usepackage{lipsum}

\usepackage[font={footnotesize,it}]{caption}

\definecolor{blue}{RGB}{0,0,0}
\definecolor{blue1}{RGB}{0,0,0}
\definecolor{blue2}{RGB}{0,0,0}

 \usepackage{fancyhdr}
\usepackage{ragged2e}
\fancypagestyle{ieee_notice}
{
    \setlength{\headheight}{32pt}
    \addtolength{\topmargin}{-5pt}
    \lhead{\fbox{\begin{minipage}{\textwidth}
    \noindent\footnotesize\justifying\copyright 2026 IEEE this paper is currently in submission to RA-L. Personal use of this material is permitted. Permission from IEEE must be obtained for all other uses, in any current or future media, including reprinting/republishing this material for advertising or promotional purposes, creating new collective works, for resale or redistribution to servers or lists, or reuse of any copyrighted component of this work in other works.\end{minipage}}}
    \cfoot{} 
}

\title{Deformable In-Hand Slip-Aware Tactile Sensor with Integrated Velocity Sensing, Force/Torque  and Pressure Map Estimation
}

\author{Gabriel Arslan Waltersson$^{1}$ and Yiannis Karayiannidis$^{2}$
\thanks{Manuscript received: May, 13, 2026; Revised August, 11, 2026; Accepted September, 5, 2026.}
\thanks{This paper was recommended for publication by Editor Ashis Banerjee upon evaluation of the Associate Editor and Reviewers’ comments.
This work was supported by Wallenberg AI, Autonomous Systems and Software Program (WASP) (Knut and Alice Wallenberg Foundation).}
\thanks{$^{1}$Gabriel Arslan Waltersson is  with the Department of Electrical Engineering, Chalmers University of Technology, SE-412 96 Gothenburg, Sweden 
         {\tt\small gabwal@chalmers.se}. 
 {\textit{Corresponding author.}}}
\thanks{$^{2}$Yiannis Karayiannidis is with the Department of Automatic Control, Lund University, Sweden {\tt\small  yiannis@control.lth.se}. The author is a member of the ELLIIT Strategic Research Area at Lund University. }
\thanks{Digital Object Identifier (DOI): see top of this page.}
}
\begin{document}

\maketitle
\thispagestyle{ieee_notice} 


\begin{abstract}

This paper introduces a novel tactile sensor for in-hand manipulation with slip-aware control that integrates \textcolor{blue2}{velocity and force/torque sensing with pressure map estimation} into a single device with a deformable contact pad. To the best of our knowledge, this is the first sensor to combine these sensing modalities within a single compliant structure. \textcolor{blue}{The sensor features a deformable contact surface and can robustly track both flat and curved surfaces across a wide range of diffuse surface materials.} Its performance is evaluated through a comprehensive set of experiments that highlight both its capabilities and limitations. The sensor is designed for rapid and low-cost fabrication using a combination of standard PCB manufacturing and rapid prototyping techniques.

\end{abstract}


\section{Introduction}

\IEEEPARstart{T}{actile} sensing is fundamental to in-hand manipulation, as it enables the estimation of contact states, the assessment of grasp stability, and reasoning about the attributes and manipulation affordances of an object. Equally important is compliance, which allows forces to be distributed across the contact surface and adapted to object geometry. The integration of tactile sensing and compliance further enables the controlled use of slippage, an integral component of in-hand manipulation for parallel grippers. Together, these mechanisms minimize effort while maintaining dexterity, providing essential perceptual and mechanical foundations for robust manipulation under uncertainty.

\begin{figure}
    \centering
    \includegraphics[width=0.79 \linewidth]{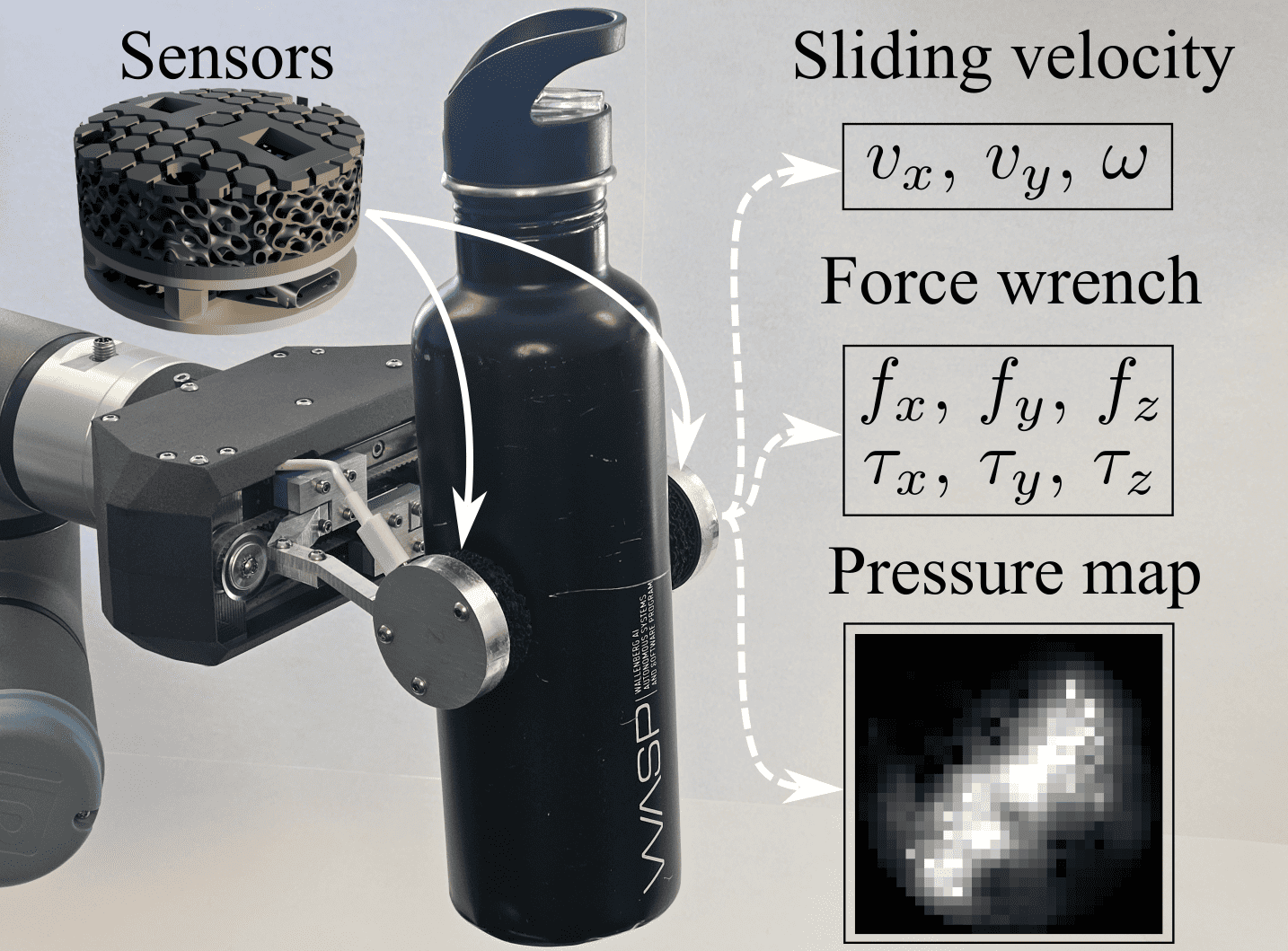}
    \caption{Tactile sensor for in-hand slip-aware object manipulation, mounted on a custom gripper.}
    \label{fig:overview}
    \vspace{-0.6cm}
\end{figure}

In-hand manipulation with parallel grippers is governed by the interaction between the contact pad and the object. These interactions depend on the friction and pressure distribution across the contact surface \cite{waltersson2024planar}, and knowledge of both can be used to synthesize controllers that regulate the object's pose relative to the hand, reason about contact states, and assess grasp stability. Friction identification is a well-studied topic across physics, materials science, robotics, and control, and numerous models have been developed to describe it under well-defined conditions. However, their practical applicability to robotic manipulation is limited. Robotic systems must interact with a wide variety of objects, often subject to wear, surface variability, and dynamically changing contact conditions. Many frictional properties are revealed only through interaction, requiring relative motion and multimodal sensing for meaningful estimation. Such sensing must include velocities, pressure distributions, torques, and both normal and tangential contact forces, integrated in real time during manipulation. This underscores the demand for tactile sensors that go beyond simple contact-detection; they must  provide rich, dynamic information about contact states and the progression of the grasp. This paper introduces a new tactile sensor (see Fig. \ref{fig:overview}) designed for in-hand manipulation with parallel grippers. This sensor measures planar sliding velocities, 6-DoF Forces/Torques (F/T), and the contact pressure distribution. \footnote{The files and code developed for this paper can be found at \url{https://github.com/Gabrieleenx/Deformable_Tactile_Sensor}}

There are many different types of tactile sensors, ranging from flexible, thin capacitive sensors \cite{dobrzynska2012polymer} capable of measuring 3D forces to visuotactile sensors \cite{lambeta2020digit} that provide high spatial resolution. Numerous reviews survey the state of tactile sensing \cite{Chi2018review, Lyu2025review, Chen2018review, Dahiya2010review, YOUSEF2011review, KAPPASSOV2015review}.  Their fabrication methods span advanced deposition techniques \cite{dobrzynska2012polymer} to readily accessible 3D printing and rapid prototyping approaches \cite{pattabiraman2025eflesh}, where \cite{11159583} did a comparison study of the different sensing modalities. However, few of these sensors focus on facilitating controlled, slip-aware in-hand manipulation, leaving a gap in the literature on tactile sensing for in-hand manipulation with planar slippage. Sliding velocity has been inferred using partially observable estimators \cite{Costanzo2023}, though these are typically limited to primary rotational slippage. Other works estimate linear sliding velocity using oscillatory features \cite{Gloumakov2024_frequency, Chen}, but these features are not consistent across materials and surfaces. Huh \emph{et al.} \cite{Huh2020Reconfigurable} designed a sensor for in-hand manipulation with dynamically reconfigurable sampling of nib deflections, allowing either fast or high-resolution data acquisition depending on the task. In \cite{vina2016slip}, tactile sensing was combined with external visual tracking for rotational slip control. Recent fiber-optic tactile sensors have also been explored for in-hand manipulation \cite{TRIPICCHIO2023fiber, Qian2018fiber, Lu2022fiber}, primarily tracking sliding velocity from changes in the pressure distribution. Different geometrical features \textcolor{blue2}{have} been used to estimate sliding velocity and position \cite{t_mech_1}.

Prior work \cite{waltersson2024inhand} presented an approach for obtaining in-hand velocity and force measurements by combining optical mouse sensors with a commercial F/T sensor embedded in each fingertip. The design presented here addresses several limitations of that approach and extends their capabilities to curved objects. The main contributions are:
\begin{itemize}
    \item A tactile sensor featuring a deformable contact pad that enables optical measurement of planar sliding velocity, estimation of full 6-DoF F/T, and reconstruction of contact pressure distribution through pad deformation using Hall-effect sensors. Other earlier uses of optical mouse sensors focused primarily on slip detection or slip avoidance \cite{Maldonado2012Mouse, Höver2010Mouse, Sani2011Mouse}, here we are targeting in-hand slip-aware control for in-hand manipulation tasks. The novelty with respect to existing tactile sensors lies in the integration of multiple sensing modalities not previously combined within a single system. 
    \item Enhanced contact dynamics and increased contact torque authority enabled by the deformable contact pad, compared to the rigid design in \cite{waltersson2024inhand}, enabling smoother and more stable manipulation, and in-hand manipulation of objects with curved surfaces.
\end{itemize}

The resulting sensor extends the perceptual and manipulation capabilities of parallel grippers by functioning as an underactuated planar joint at the end-effector. It provides relative sliding velocities, full 6-DoF F/T estimates, and spatial pressure information within a single integrated design. These measurements support robust estimation of grasp state, in-hand motion, object properties, and external contacts, enabling friction-aware and compliance-based manipulation strategies. To the best of our knowledge, this combination of capabilities has not previously been realized in a single tactile sensor.

\section{\textcolor{blue}{Tactile Sensor Design}}
The tactile sensor is designed to combine multiple sensing modalities, enabling the measurement of sliding velocities, 6-DoF forces and torques, and the contact pressure map. All components are integrated into a single sensor unit, together with a microcontroller and a standard USB-C interface for both power and communication. The sensor interacts with objects through a deformable contact pad, which provides desirable contact properties and dynamics. The sensor is constructed in multiple layers and can be manufactured using rapid prototyping techniques. The overall dimensions of the sensor are 40 mm in diameter and approximately 22 mm in height in its resting configuration, \textcolor{blue2}{and} the sensor weighs 23 grams. An exploded view of the sensor is shown in Fig. \ref{fig:exploded}.

\subsection{Construction and Manufacturing}
The base of the sensor consists of two PCBs stacked with 5 mm spacers and connected using a board-to-board connector. The bottom PCB houses the USB-C interface, the ESP32-S3 microcontroller, and the voltage regulators. 
The spacers serve a dual purpose: attaching the contact pad and providing mounting points for installing the sensor on a gripper's fingers. The top PCB contains all sensing elements: 13 Hall-effect sensors (MLX90393) and two optical mouse sensor units (PAW3335DB). It includes two cutouts to provide additional travel for the mouse sensors during contact pad deformation and to maintain a compact design. Two separate SPI buses are used to communicate with the Hall-effect sensors and the optical sensors, respectively. The PCBs are ordered with machine-assembled components; only the optical sensor units \textcolor{blue2}{have} to be soldered by hand.

\begin{figure}[!t]
    \centering
    \makebox[\linewidth][c]{%
    \subfloat[]{%
        \includegraphics[width=0.44\linewidth
        ]{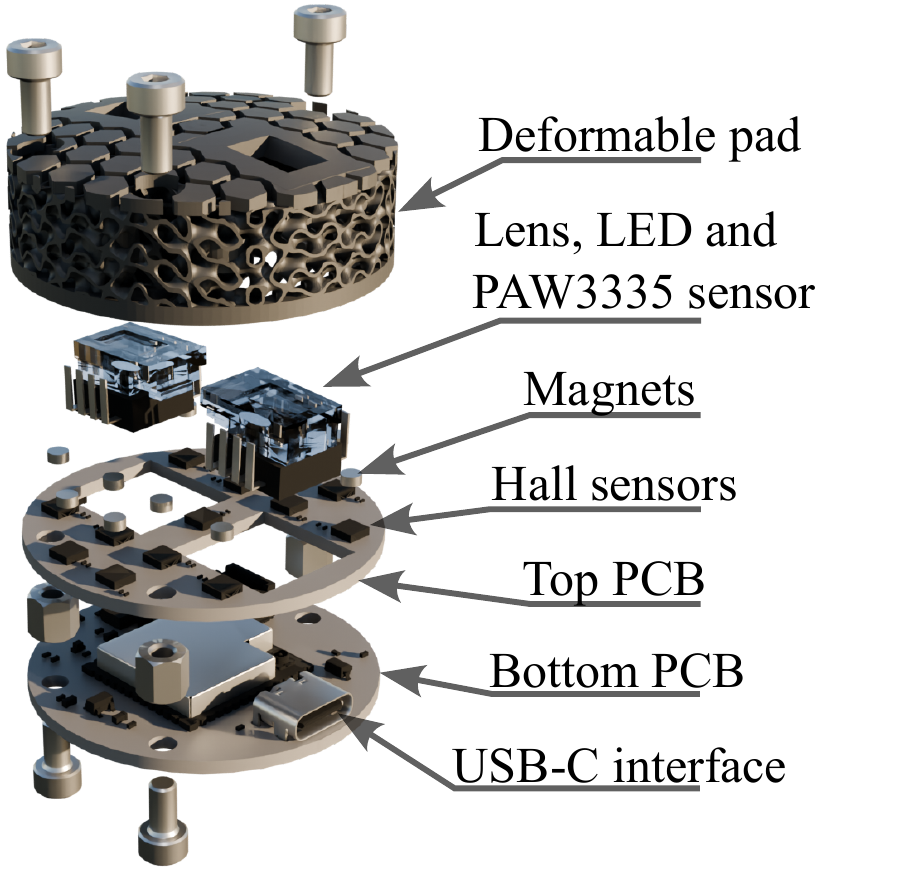}%
        \label{fig:exploded}%
    }%
    \subfloat[]{%
        \includegraphics[width=0.31\linewidth
        ]{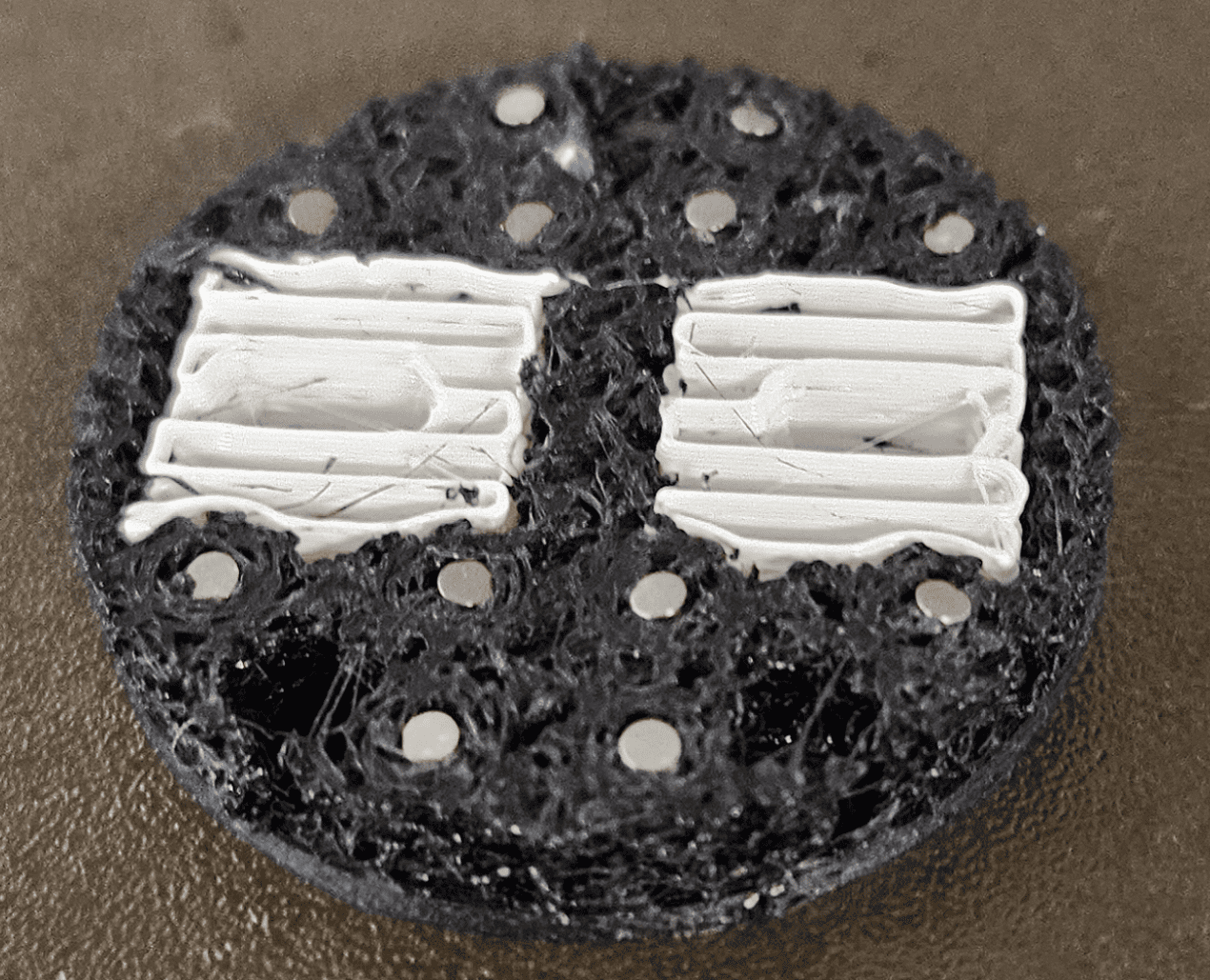}%
        \label{fig:magnets}%
    }%
    \subfloat[]{%
        \includegraphics[
            width=0.25\linewidth
        ]{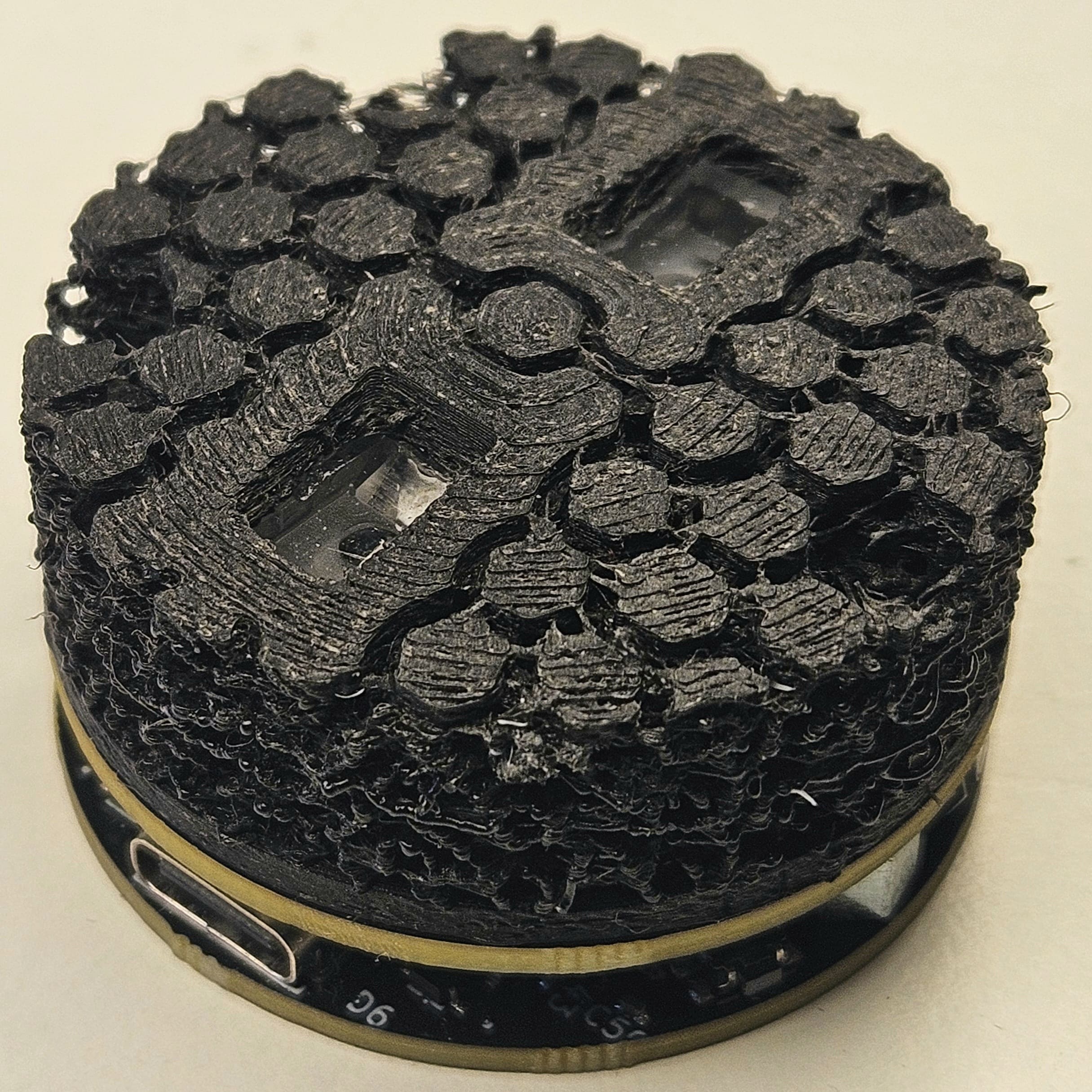}%
        \label{fig:sensor_assembled}%
    }%
    }

    \caption{Manufacturing and the assembled sensor. (a) Exploded view of the sensor. (b) Magnets placed inside of the deformable contact pad. (c) The assembled sensor.}
    \label{fig:mouse_sensor_real}
    \vspace{-0.5cm}
\end{figure}

The deformable contact pad is 3D-printed with a gyroid minimal-surface and a hexagonal top pattern. The gyroid structure provides approximately uniform deformation properties in all directions, while the hexagonal surface pattern improves wear resistance while still allowing local deformation to propagate. The contact pad includes cutouts and mounting features for the two optical sensors, which follow the deformation motion. The pad is printed as a single piece using TPU 85A, with PLA used as support material in an FDM printer. During printing, 12 magnets (N35, 2 × 1 mm) are inserted into the pad (Fig. \ref{fig:magnets}). A small amount of glue is applied beneath each magnet before insertion, after which printing continues. Although the system uses 13 Hall-effect sensors, only 12 magnets are embedded to allow compensation for non-local external magnetic disturbances such as the Earth’s magnetic field.

\begin{figure}[!t]
    \centering

    \subfloat[]{
        \includegraphics[
            trim=50 0 0 0,
            clip,
            width=0.4\columnwidth
        ]{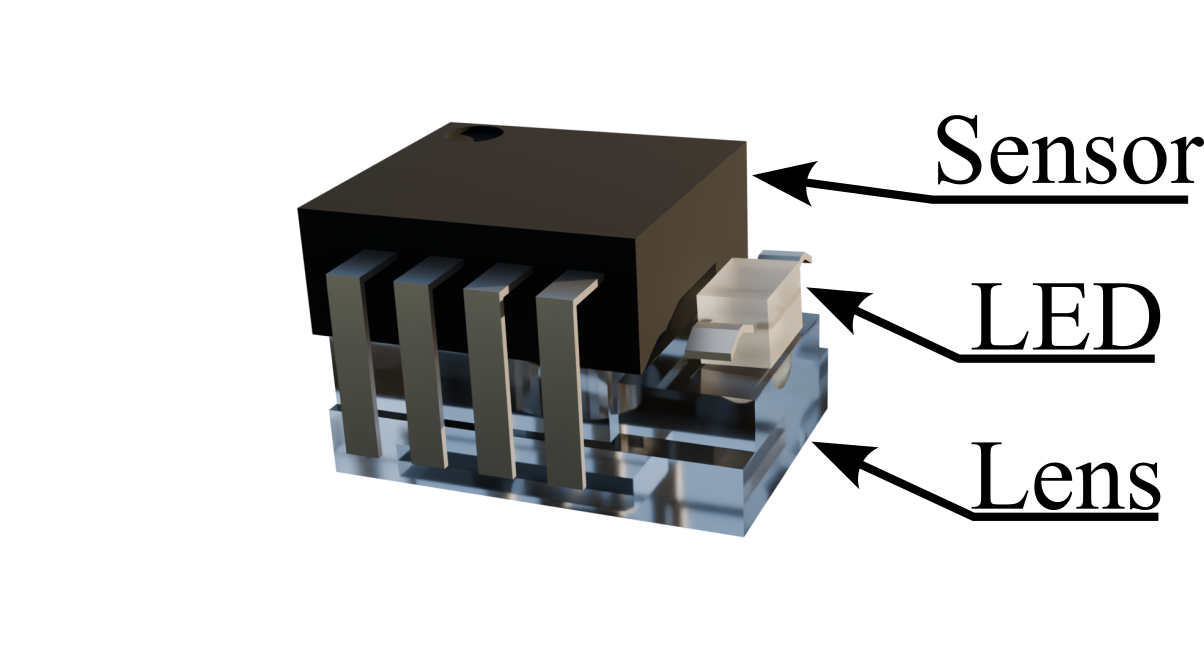}
        \label{fig:mouse_sensor_unit}
    }
    \subfloat[]{
        \includegraphics[
            trim=800 300 600 600,
            clip,
            width=0.4\columnwidth
        ]{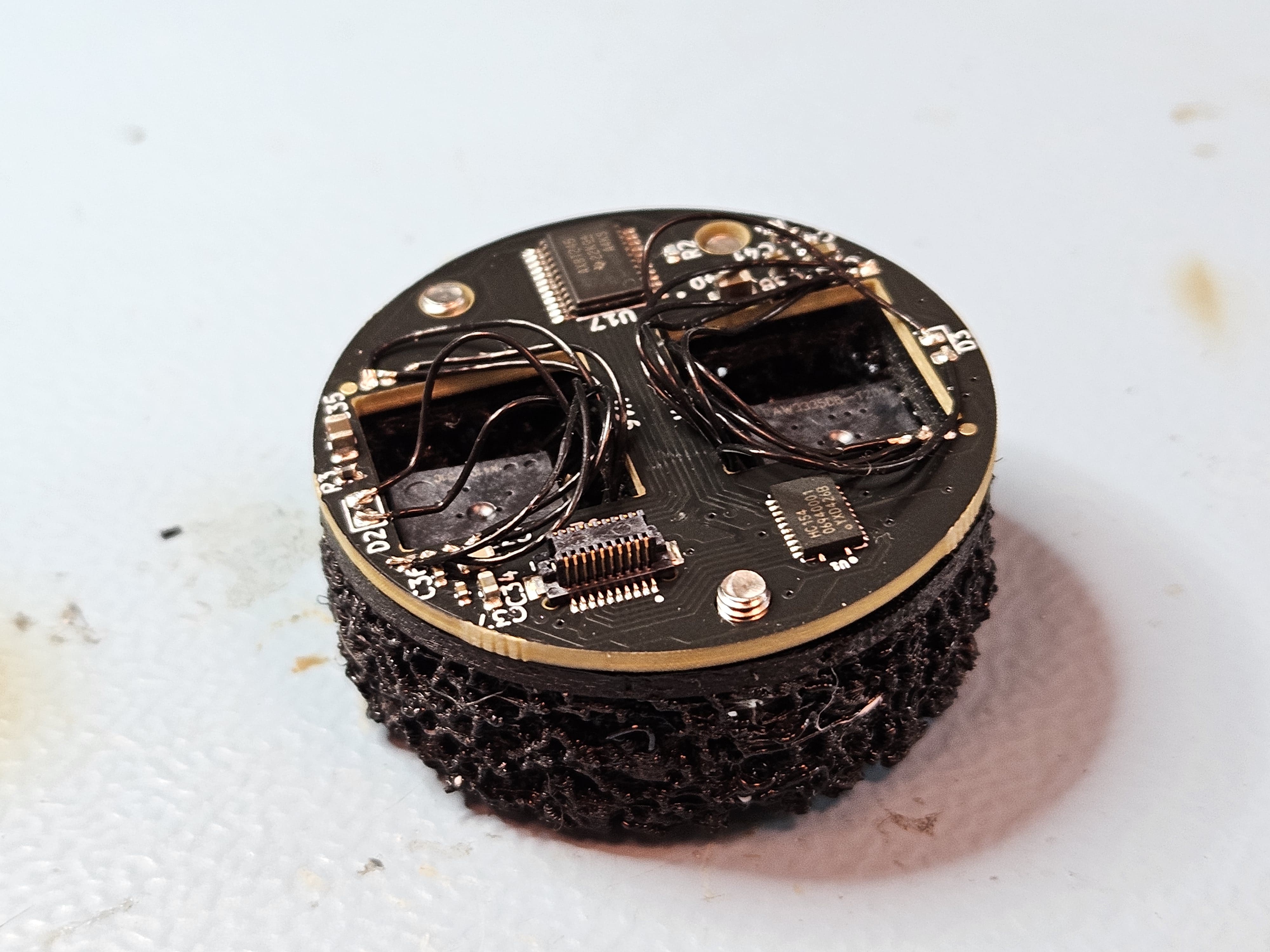}
        \label{fig:mouse_sensor_unit_soldered}
    }

    \caption{Construction and assembly of the computer mouse sensor units. (a) Computer mouse sensor unit. (b) Two mouse sensor units inserted into the contact pad. }
    \label{fig:mouse_sensor}
    \vspace{-0.5cm}
\end{figure}

Each optical mouse sensor unit measures $x$- and $y$-displacement, which are used to compute the linear sliding velocities $v_x$, $v_y$, and the rotational velocity $\omega$ by combining the two sensor readings. Each unit consists of the PAW3335DB sensor, set to 4900 counts per inch and a cutoff distance of 2 mm, an IR LED, and a lens (Fig. \ref{fig:mouse_sensor_unit}). The lens, taken from the PAW3335DB sample kit, is cut to reduce its footprint. Both the LED and the sensor are glued directly to the lens. Thin and flexible 30 AWG, 49-strand wires are used to solder the sensor units to the underside of the top PCB (Fig. \ref{fig:mouse_sensor_unit_soldered}). The completed sensor units are then inserted into press-fit slots in the contact pad, with a small amount of glue added to prevent loosening.

Together, these components enable a sensor capable of tracking F/T, pressure distributions, and sliding velocities of objects in-hand without relying on external vision. All Hall-effect and mouse sensor readings are streamed at 250 Hz over USB. The fully assembled sensor is shown in Fig. \ref{fig:sensor_assembled}.

\section{\textcolor{blue}{Sensor Calibration and Mapping}}

The raw measurements from the tactile sensor must be mapped into the desired measurement space. This includes mapping the Hall-effect measurements to F/T, pressure distribution, and the magnetic coupling with the 13th Hall sensor, as well as calibrating the poses of the optical mouse sensors.

\subsection{Hall-Effect Mapping}

\begin{figure}
    \vspace{0.3cm}
    \centering
    \includegraphics[width=0.85\linewidth]{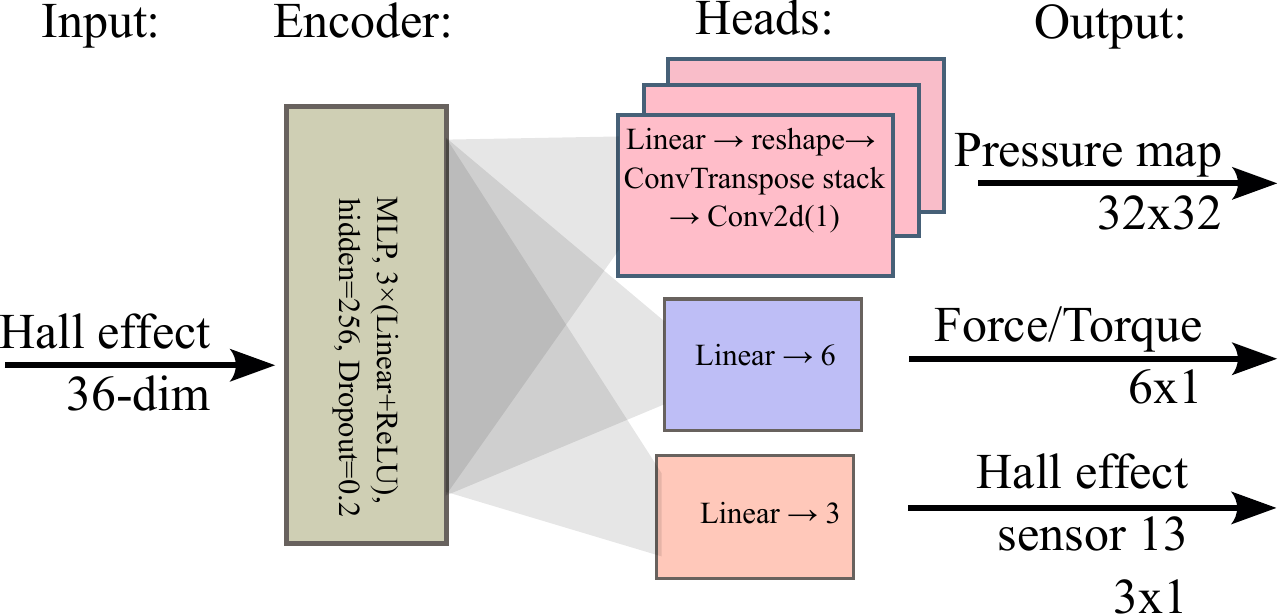}
    \caption{Neural network architecture for mapping Hall-effect measurements to pressure maps, F/T, and magnetic disturbance compensation.}
    \label{fig:network}
    \vspace{-0.5cm}
\end{figure}

The 12 Hall-effect sensors with magnets above them are mapped to three quantities: the 6-DoF F/T measurements, the contact pressure distribution, and the magnetic cross-influence on the 13th Hall-effect sensor, which does not have a magnet directly above it. Neural-network-based mapping from Hall-effect sensor measurements to force and contact location has been investigated previously \cite{pattabiraman2025eflesh}. Since the contribution of this work lies beyond the mapping methodology itself, we adopt a simple neural network architecture composed of a shared encoder and three task-specific output heads (Fig. \ref{fig:network}).

A shared encoder learns a latent representation of the underlying deformation state of the sensor, while the output heads decode task-specific quantities from this latent representation \cite{ruder2017overview}. The encoder is a three-layer MLP with 256 hidden units per layer, ReLU activations, and dropout (0.2) applied to the first two layers during training. From the shared latent representation, three heads branch out:
(i) a deconvolutional decoder producing a normalized $32 \times 32$ pressure map,
(ii) a fully connected head predicting the 6 F/T values, and
(iii) a fully connected head predicting the 3-axis magnetic signal of the 13th Hall-effect sensor.
The 13th sensor is used to compensate for external magnetic disturbances, and the network learns how the 12 embedded magnets influence it.

Each Hall-effect measurement is first filtered using an exponential running average, where the filtered value at time step $k$ is given by:
\begin{equation}
\hat{m}_{i,j}(k) = \alpha_m (m_{i,j}(k) - b_{i,j}) + (1 - \alpha_m)\hat{m}_{i,j}(k-1),
\end{equation}
where $m_{i,j}$ is the magnetic reading from sensor $i$ and axis $j \in {x,y,z}$, $b_{i,j}$ is a bias term used to re-zero the sensor when not in contact and $\alpha_m$ is the forgetting factor\textcolor{blue2}{, balancing noise attenuation and response delay.}

During deployment, the network input is corrected for external disturbances as:
\begin{equation}
\hat{\bar{m}}_{i,j}(k) = \hat{m}_{i,j}(k) - \gamma_m \delta m_j(k-1),
\end{equation}
where $0 \leq \gamma_m \leq 1$ is a scalar gain and
\begin{equation}
\delta m_j(k) = \hat{m}_{13,j}(k) - f_{h13,j}(\hat{\bar{m}}(k))
\end{equation}
represents the estimated external magnetic disturbance computed as the difference between the measurement of the 13th Hall-effect sensor and the predicted magnetic influence $f_{h13,j}$ from the other 12 magnets. Because part of the network output is fed back into its input, $\gamma_m$ is used to limit the influence of this correction and reduce the risk of feedback instability.

\subsubsection{Data collection}
\begin{figure}[!t]
    \centering

    \subfloat[]{
        \includegraphics[width=0.30\columnwidth]{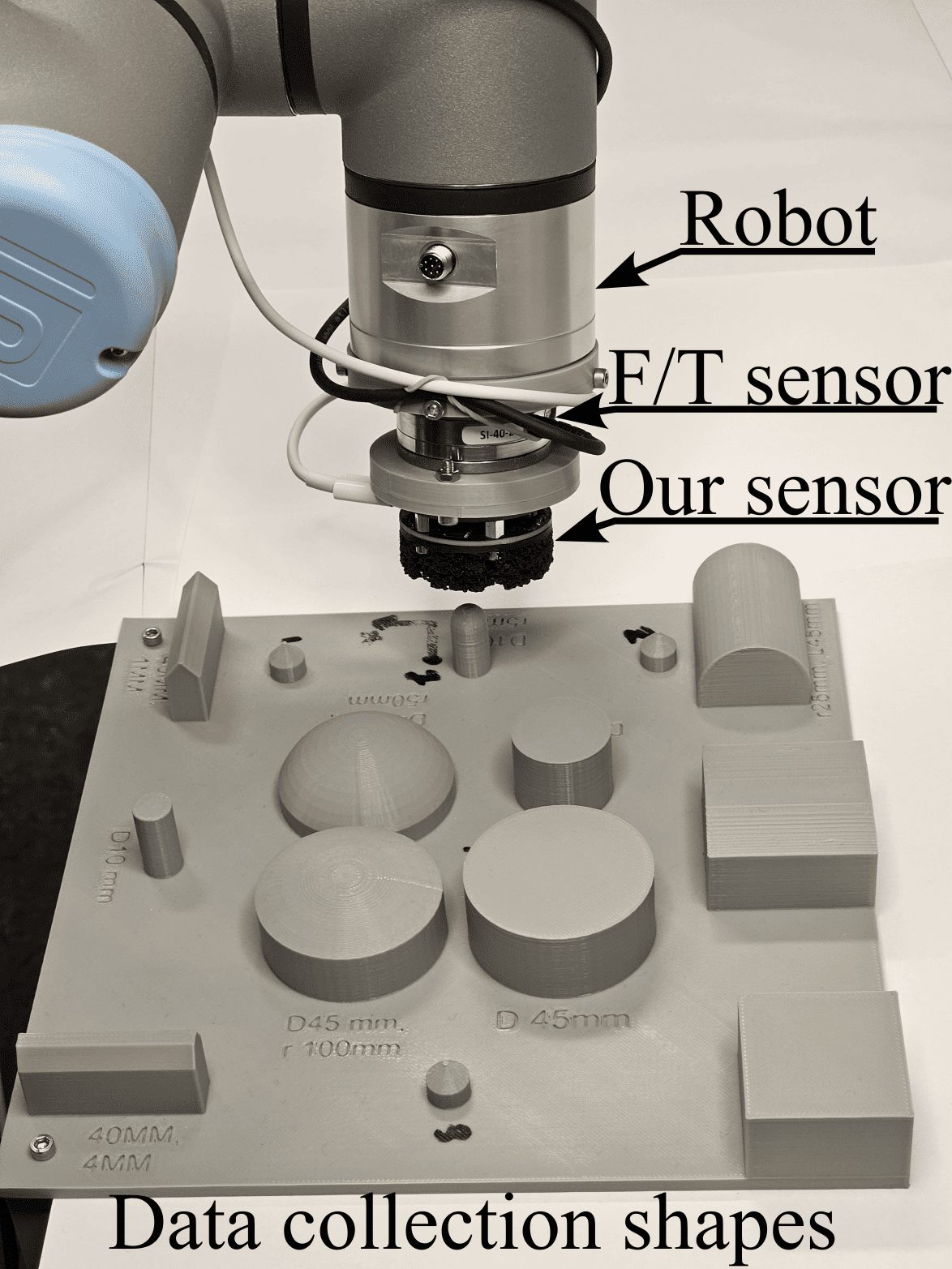}
        \label{fig:tactile_gym}
    }
    \subfloat[]{
        \includegraphics[width=0.30\columnwidth]{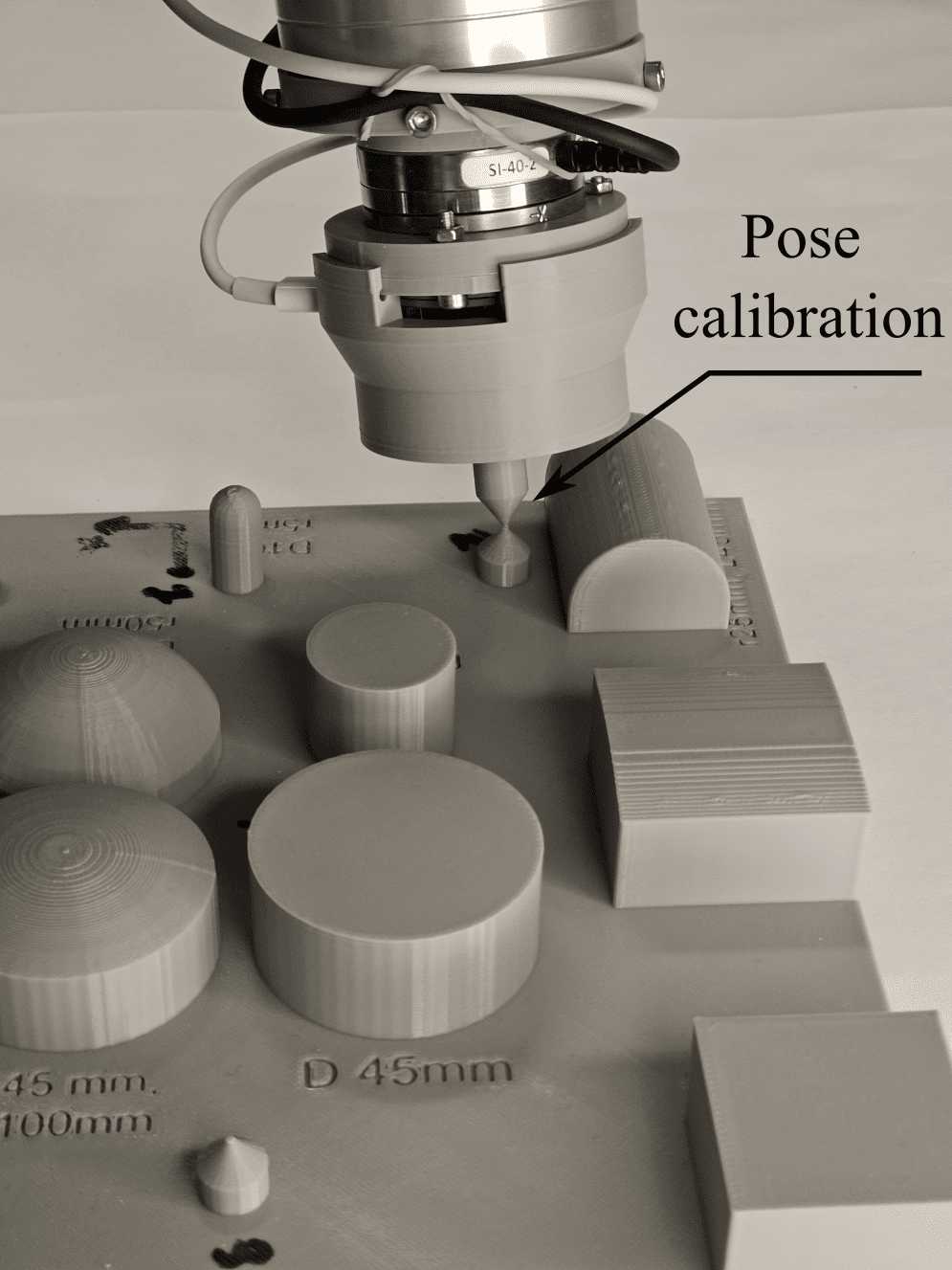}
        \label{fig:calibration_pose}
    }
    \subfloat[]{
        \includegraphics[width=0.30\columnwidth]{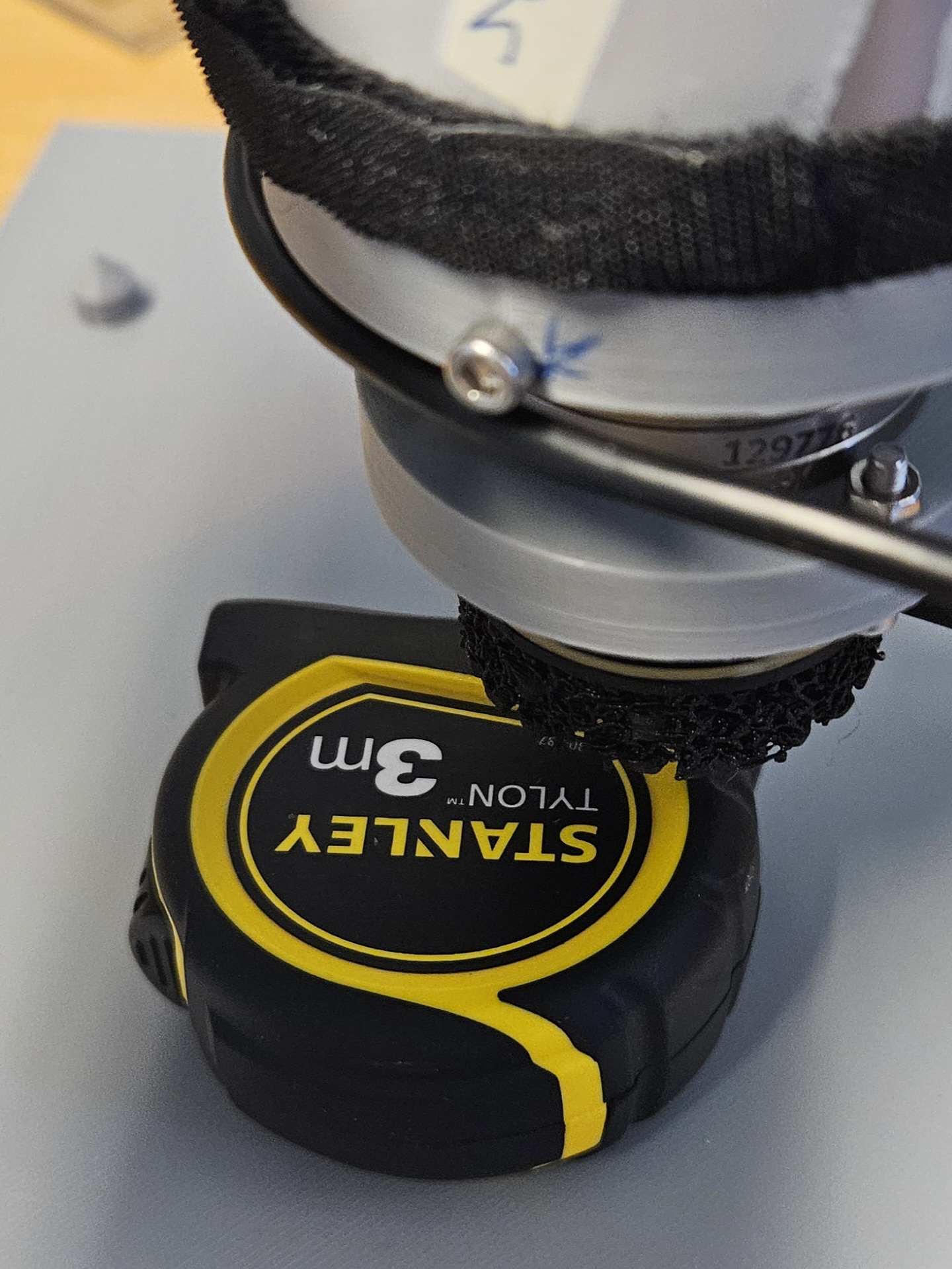}
        \label{fig:unseen_object}
    }

    \caption{Tactile exploration and data collection for learned sensor mapping. (a) Data collection with a UR3e. (b) Calibration of pose. \textcolor{blue}{(c) Unseen object.}}
    \label{fig:tactile_collection}
    \vspace{-0.5cm}
\end{figure}

Data collection is performed using a robot equipped with the tactile sensor and a wrist-mounted F/T sensor. The robot interacts with a set of shapes protruding from an exploration surface (Fig. \ref{fig:tactile_gym}); the objective of the exploration phase is to sample the full 6-DoF force space over diverse contact locations and contact distributions. First, the pose of the exploration object is estimated by recording three known positions using a calibration tool tip (Fig. \ref{fig:calibration_pose}). The tool tip is then removed, and the tactile sensor is used to explore the shapes.

The data collection procedure begins by touching a flat, cylindrical surface (see Fig. \ref{fig:tactile_gym}) to calibrate the contact onset height. To collect data covering normal forces, tangential forces, and torques about the $z$-axis, the sensor is then pressed against the flat surface using a predetermined set of normal force steps $[2, 4, 8, 12, 16, 20]$ $N$. For each normal force step, the sensor is first rotated about the $z$-axis in both clockwise and counterclockwise directions to generate torque data. The sensor is then translated 4 mm in a random $x,y$ direction and returned to the initial position to collect tangential force data. This random tangential motion is repeated three times for each force level.

Following this calibration sequence, each object shape (Fig. \ref{fig:tactile_gym}) is poked 30 times using a randomized exploration policy. For each trial, a 6-DoF target pose is sampled uniformly within a predefined workspace. The exploration policy first samples a random 
$x,y$ contact location from a bounded region specific to each shape, after which the sensor is lowered to the previously calibrated contact onset height. From this position, a random 6-DoF delta motion is sampled uniformly from predefined bounds, allowing the sensor to interact with each object at random locations and orientations to obtain coverage of forces, torques, and contact pressure distributions.

\subsubsection{Data processing}

From the collected experimental data, we construct a dataset comprising 6-DoF F/T measurements, Hall-effect sensor readings, and estimated contact pressure distributions. The contact pressure distribution is \textcolor{blue}{approximated} by computing the relative pose of the sensor with respect to the object, applying it to a digital mesh model, and evaluating the overlap between the object and the sensor geometry. The overlap thickness is then projected onto the sensor frame to form a 60$\times$60 image, which is normalized \textcolor{blue}{and down-sampled to 32$\times$32} to produce the estimated pressure map. Each dataset entry consists of the normalized pressure map, the 13 Hall-effect readings, and the F/T measurements expressed in the tactile sensor frame. To reduce computational cost and maintain class balance, we subsample the data by selecting every 50th contact sample and every 800th non-contact sample.

\subsubsection{Training}
The network is trained using the loss:
\begin{equation}
L = L_{\texttt{map}} + 5 L_{\texttt{F/T}} + L_{\texttt{h13}},
\end{equation}
where $L_{\texttt{map}}$, $L_{\texttt{F/T}}$, and $L_{\texttt{h13}}$ denote mean squared error (MSE) losses for the pressure map, F/T, and 13th Hall-effect predictions, respectively. For $L_{\texttt{map}}$, a masked MSE is used so that only the circular region corresponding to the active sensor surface contributes to the loss. The scalar multiplier applied to $L_{\texttt{F/T}}$ balances the relative magnitudes of the loss terms. Training is performed for 25 epochs on a dataset of 15184 samples, using a batch size of 512 and a learning rate of $1\times 10^{-4}$ with the Adam optimizer. An independently collected validation set of 3051 samples is used \textcolor{blue2}{to} test the model.

\subsection{Velocity sensor}

The optical sensors are embedded in the deformable contact pad. The precise location where they track the surface is not directly known. Instead we mount the sensor on a robot and perform predetermined sliding motions, the data from this experiment \textcolor{blue2}{are} then used to estimate the pose of the two optical sensors. The planar sliding velocity of the entire sensor    
$
    \mathbf{v}^{s} = 
\begin{bmatrix}
v_x^{s}  \;
v_y^{s}  \;
\omega^{s}
\end{bmatrix}^\top
$
is  mapped to the optical sensors' individual velocities   $\mathbf{v}^{mi}$ as follows:
\begin{equation}
    \begin{bmatrix}
        \mathbf{v}^{m1} \\
        \mathbf{v}^{m2}
    \end{bmatrix} =\mathbf{H}\mathbf{v}^{s} \;\textrm{with}\;\mathbf{H}=\begin{bmatrix}
        R_z^\top(\theta_1) & R_z^\top(\theta_1) p_{\perp}(\mathbf{r}_1) \\
        R_z^\top(\theta_2) & R_z^\top(\theta_2) p_{\perp}(\mathbf{r}_2)
    \end{bmatrix} 
\end{equation}
where $\theta_i$ is the orientation of the optical sensor $i$ in the sensor frame, $R_z(\theta_i)\in SO(2)$, $\mathbf{r}_i= 
\begin{bmatrix}
r_{xi}  \;
r_{yi}  
\end{bmatrix}^\top
$ is the local position vector of the optical sensor $i$ and $p_{\perp}(\mathbf{r}_i)= 
\begin{bmatrix}
-r_{yi}  \;
r_{xi}
\end{bmatrix}^\top
$ its orthogonal complement. 
Finally, from the combined velocity measurements from each of the optical sensors the planar sliding velocity of the sensor can be obtained by:
\begin{equation}
    \mathbf{v}^{s} = \mathbf{H}^{-1} \begin{bmatrix}
        \mathbf{v}^{m1} \\
        \mathbf{v}^{m2}
    \end{bmatrix}
\end{equation}
The calibration procedure is performed in two steps. First a predetermined linear motion is performed on a flat surface. From this data the (Counts per mm) CPMM can be estimated and both sensor orientations  from: 
\begin{equation}
    \theta_i = \arctan2(\Delta y^{s}, \Delta x^{s}) -  \arctan2(\Delta y^{mi}, \Delta x^{mi})
\end{equation}
where $\Delta y^{s}$ and $\Delta x^{s}$ are the displacements of the sensor along $y$ and $x$ directions respectively, and $\Delta y^{mi}$ and $ \Delta x^{mi}$ are the measured displacements by the optical sensor $i$.
Second a rotation \textcolor{blue}{$\Delta \theta$} is performed, and using the CPMM and orientation estimation in the previous step, the 2D pose of the optical sensors is estimated with \textcolor{blue2}{regard} to the tactile sensor frame. \textcolor{blue}{Each optical sensor's measurements are transformed to align with the $x$- and $y$-axes of the tactile sensor frame. Each optical sensor then measures the displacements $\Delta y^{s}_{mi}$ and $\Delta x^{s}_{mi}$, from which the radial direction $\phi_i$ of the optical sensor can be calculated as:}
\begin{equation}
    \phi_i = \arctan2(\Delta y^{s}_{mi}, \Delta x^{s}_{mi})  + \pi/2. 
\end{equation}
Finally the distance along that \textcolor{blue}{radial direction} is estimated by calculating the arc length $l_i$ traveled and the radius $\xi_i = l_i/\Delta \theta$. The final poses of the \textcolor{blue2}{sensors are} calculated by $x_i = \xi_i \cos (\mathcolor{blue}{\phi_i})$ and $y_i = \xi_i \sin (\mathcolor{blue}{\phi_i})$ for the optical sensor $i$. The results from the calibration can be seen in Table \ref{tab:optical_calibration}.

\begin{table}[htbp]
\vspace{0.3cm}
    \centering
    \begin{tabular}{c|c|c|c|c}
    \hline
        Sensor & CPMM & $\theta$ & x & y  \\ \hline
         1 & 216.3 & 1.52 [rad] & -7.11 [mm] & 0.36 [mm]  \\ \hline
         2 & 212.7 & -1.56 [rad] & 8.41 [mm] & -0.17 [mm] \\ \hline
    \end{tabular}
    \caption{Calibration of optical sensors}
    \label{tab:optical_calibration}
    \vspace{-0.5cm}
\end{table}

\section{Results}
To clearly demonstrate and validate the sensor’s capabilities, a series of experiments are performed examining: (a) the sensor’s ability to measure F/T, (b) hysteresis and sensor drift over time and repeated contact, and the sensor’s ability to compensate for external magnetic fields, (c) the ability to estimate a pressure map, (d) planar velocity tracking over multiple surfaces, and (e) in-hand slip-aware control. All sensor data are streamed at 250 Hz. The neural network inference time is 1.0 ms when executed on a single CPU core (Intel Core i7-1185G7), and the total computational time, including filtering and inference, is 1.3 ms. \textcolor{blue2}{The end-to-end latency, from sensor acquisition to filtered estimates, is upper-bounded by 6.3 ms, excluding the exponential filter’s temporal response.}

\subsection{Force Torque Sensing}

\textcolor{blue}{The sensor is evaluated on a validation set collected separately from the training dataset. All experiments use $\alpha_m = 0.1$ and $\gamma_m = 0.95$. Furthermore, the sensor is evaluated on an unseen object (Fig.~\ref{fig:unseen_object}), which was poked 30 times using the same exploration procedure. Table~\ref{tab:force_torque} reports the mean error ($\mu$), standard deviation of the error ($\sigma$), normalized root mean square error (NRMSE), and the tested range for each force and torque component. The NRMSE is computed as $\mathrm{NRMSE} = \mathrm{RMSE}/R$, where $R = y_{\max} - y_{\min}$ is the range of the validation dataset; the same range is also used to normalize the unseen-object evaluation. The sensor is additionally evaluated using only samples where contact occurs, as shown in Table~\ref{tab:force_torque_in_contact}.}

\textcolor{blue}{The results show that, for objects included in the training set, the sensor achieves an NRMSE of approximately $3$--$4\%$ while in contact. For the unseen object, the estimation accuracy is comparable for most components, although the NRMSE of the normal force increases. For comparison with \cite{yuan2017gelsight}, the NRMSE was computed from the reported RMSE values and measurement ranges, yielding NRMSE values of: ($f_x$): 3.3\%; ($f_y$): 3.7\%; ($f_z$): 4.9\%; and ($t_z$): 0.8\%.
These values are of a similar magnitude to those obtained in this work, despite the different experimental setups. Nevertheless, this sensor is intended to operate as part of a multimodal system, where force measurements are complemented by additional sensing modalities.}

\begin{figure}[tbp]
    \centering
    \includegraphics[width=0.7\linewidth]{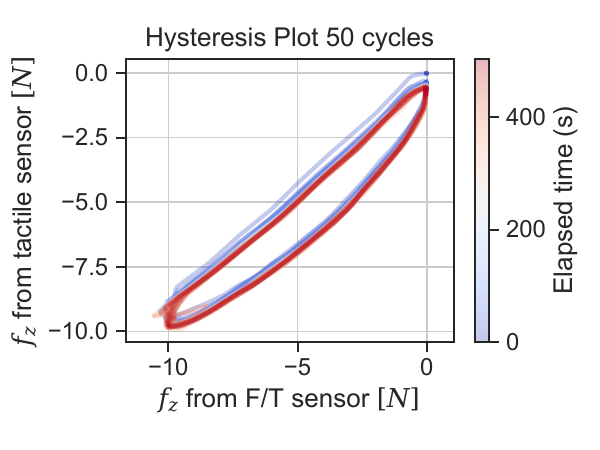}
    \caption{Hysteresis test with 50 cycles.}
    \label{fig:hysteresis}
    \vspace{-0.6cm}
\end{figure}

\begin{table}[b]
    \centering
    \resizebox{\columnwidth}{!}{%
    \begin{tabular}{c|c|c|c|c|c}
        \hline
        Type & $\mu$ & $\sigma$ & NRMSE (\%) & Unseen (\%) & Range \\ \hline
        $f_x$ [N] & 1.091e-01 & 3.831e-01 & 2.54 & 1.45& [-7.053, 8.639] \\ \hline
        $f_y$ [N] & 1.388e-02 & 2.925e-01 & 2.10 & 1.19& [-6.384, 7.534] \\ \hline
        $f_z$ [N] & 2.033e-02 & 5.755e-01 & 2.84 & 3.44& [-20.185, 0.098] \\ \hline
        $t_x$ [Nm]& -7.043e-04 & 8.560e-03 & 2.01 &1.51 &[-0.220, 0.208] \\ \hline
        $t_y$ [Nm]& 3.911e-03 & 1.152e-02 & 2.98 &1.48  &[-0.193, 0.215] \\ \hline
        $t_z$ [Nm]& -1.247e-04 & 5.756e-03 & 2.45 &  1.43& [-0.118, 0.117] \\ \hline
    \end{tabular}
    }
    \caption{\textcolor{blue}{Force and torque estimation error statistics. NRMSE is normalized by the tested range. Unseen (\%) is the NRMSE for the unseen object.}}
    \label{tab:force_torque}
\end{table}

\begin{table}[tbp]
    \centering
    \resizebox{\columnwidth}{!}{%
    \begin{tabular}{c|c|c|c|c|c}
        \hline
        Type & $\mu$ & $\sigma$ & NRMSE (\%) & Unseen (\%) & Range \\ \hline
        $f_x$ [N] & 2.332e-01 & 5.166e-01 & 3.61 & 2.85 &[-7.053, 8.639] \\ \hline
        $f_y$ [N] & 3.466e-02 & 4.115e-01 & 2.97 & 2.35 &[-6.384, 7.534] \\ \hline
        $f_z$ [N] & 8.439e-02 & 8.070e-01 & 4.04 & 7.01 &[-20.185, -0.101] \\ \hline
        $t_x$ [Nm]& -1.777e-03 & 1.203e-02 & 2.84 &  3.05&[-0.220, 0.208] \\ \hline
        $t_y$ [Nm]& 7.966e-03 & 1.548e-02 & 4.26 & 3.01& [-0.193, 0.215] \\ \hline
        $t_z$ [Nm]& -2.083e-04 & 8.229e-03 & 3.50 & 2.85& [-0.118, 0.117] \\ \hline
    \end{tabular}
    }
    \caption{\textcolor{blue}{Force and torque estimation error statistics in contact only $|f_z|>0.1$ (1482 data points). NRMSE is normalized by the tested range. Unseen (\%) is the NRMSE for the unseen object.}}
    \label{tab:force_torque_in_contact}
\end{table}

To further evaluate the F/T estimation, three known effects associated with this type of sensor are investigated: hysteresis, drift, and sensitivity to external magnetic fields. Hysteresis is tested by repeatedly bringing the sensor into contact with a flat plane for 50 cycles, each cycle lasting 10 seconds, and targeting a normal force of 10 N. The measured force is compared to that of a commercial F/T sensor and plotted in Fig. \ref{fig:hysteresis}.
The hysteresis test shows the expected viscoelastic behavior of the contact pad, where the force measurement depends on the loading direction\textcolor{blue2}{, while the F/T channel is primarily limited by the mechanical bandwidth of the contact pad rather than that of the electronics.} No significant drift is observed during this test.

\begin{figure}
    \centering
    \includegraphics[width=0.8\linewidth]{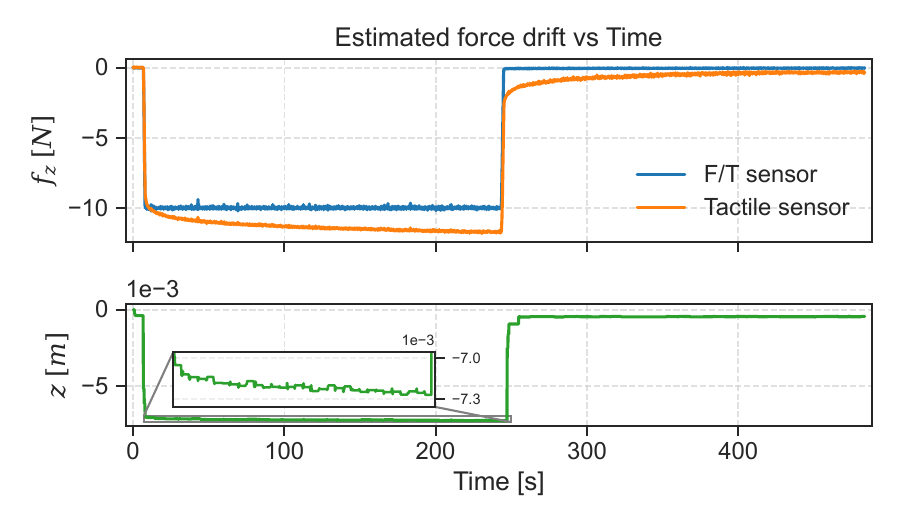}
    \caption{\textcolor{blue2}{Sensor kept at 10 N normal force for 240 seconds.} }
    \label{fig:drift}
    \vspace{-0.6cm}
\end{figure}

Drift is further evaluated by maintaining prolonged contact at 10 N\textcolor{blue}{, controlled by the commercial F/T sensor,} for 240 s, followed by an equally long de-loading phase. The results are shown in Fig. \ref{fig:drift}.
The results indicate that the sensor experiences drift during prolonged contact, which is attributed to slow deformation of the compliant contact pad under sustained pressure.
The deformable pad undergoes an additional deformation of approximately 0.3 mm, \textcolor{blue}{estimated from the robot forward kinematics,} beyond the initial compression during the contact phase, corresponding to an apparent increase of approximately 2 N in estimated normal force, despite the applied force remaining constant. A similar effect is observed during unloading, where the sensor initially recovers quickly before slowly relaxing the remaining deformation. \textcolor{blue}{These long-term viscoelastic effects are not explicitly captured during data collection.}

\begin{figure}[htbp]
    \vspace{0.25cm}
    \centering
    \includegraphics[width=0.8\linewidth]{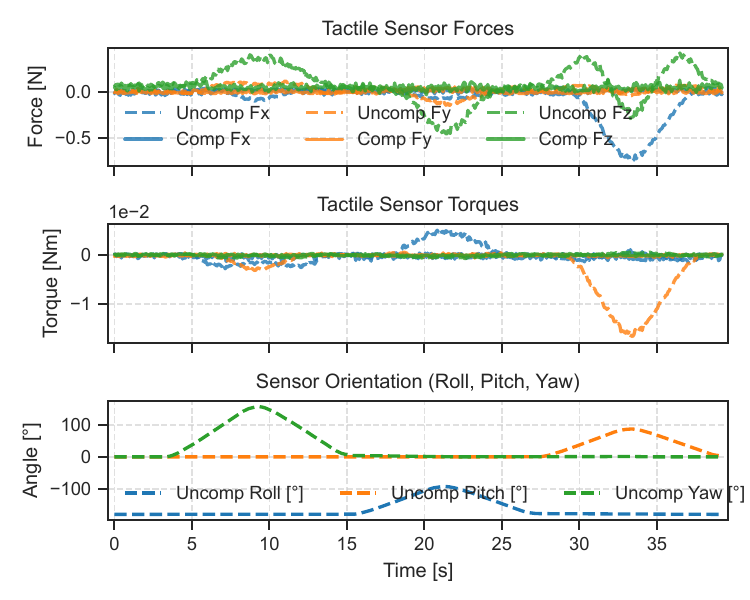}
    \caption{Sensor compensation for external magnetic fields, different orientations \textcolor{blue2}{change} the relative effect of Earth's magnetic field }
    \label{fig:external_comp}
    \vspace{-0.4cm}
\end{figure}

Another limitation of Hall-effect sensors is their sensitivity to external magnetic fields \cite{earth_mag, t_mech_2}, such as the Earth’s magnetic field. This is particularly relevant when the sensor is mounted on a robotic platform that undergoes orientation changes. The central Hall-effect sensor does not have a magnet positioned above it, and cross-coupling from the surrounding magnets is estimated using a learned mapping. The output of this central (13th) Hall-effect sensor is then used to compensate for external magnetic bias. \textcolor{blue2}{No feedback instability was observed.} Estimated F/T values are recorded during changes in sensor orientation, both with and without magnetic compensation, as shown in Fig. \ref{fig:external_comp}. The results show that the Earth’s magnetic field can introduce errors of up to 0.7 N. With compensation enabled, no noticeable error due to orientation changes is observed. \textcolor{blue2}{The compensation method assumes a uniform external magnetic field affecting all 13 Hall-effect sensors equally.}

\begin{figure}[htbp]
    \centering
    \includegraphics[width=0.99\linewidth]{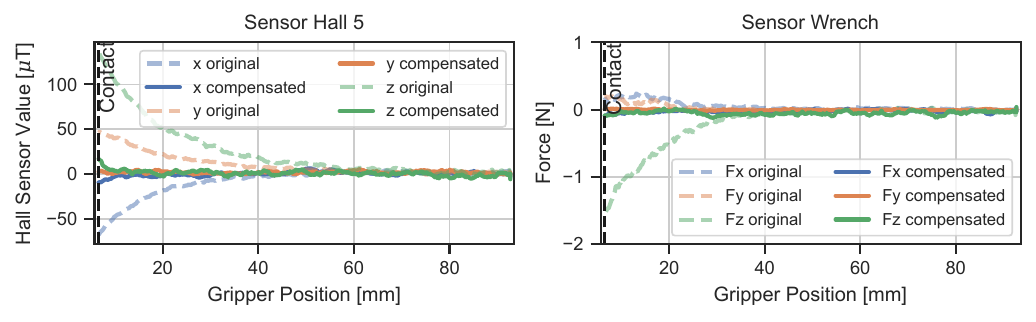}
    \caption{Cross sensor interference and compensation, \textcolor{blue2}{the left plot shows} the reading from hall sensor 5 and the right is for the estimated force wrench of the sensor. \textcolor{blue2}{The maximum (mean) residual force magnitude was $1.594$ N ($0.205$ N) before and $0.130$ N ($0.056$ N) after compensation.}}
    \label{fig:sensor_inteferance}
    \vspace{-0.5cm}
\end{figure}

These sensors are intended to be mounted on a gripper, as shown in Fig.~\ref{fig:overview}. When the two sensors are positioned close to each other, they exhibit magnetic cross-interference (Fig.~\ref{fig:sensor_inteferance}). \textcolor{blue2}{This interference was empirically modeled as an exponential decay}, $y(x) = A e^{-k(x-x_0)} + C$, where $x$ is the gripper finger position and $x_0$ is the position at which the fingers make contact. The model parameters are fitted independently for each Hall-effect sensor and measurement direction. As shown in Fig.~\ref{fig:sensor_inteferance}, the magnetic cross-interference can be effectively compensated \textcolor{blue2}{over the full grasp range}. Furthermore, electromagnetic interference from the gripper motor was evaluated by mechanically jamming the gripper fingers while allowing current to flow through the motor. No measurable interference was detected. \textcolor{blue2}{It should be noted that local magnetic disturbances may still affect the sensor.}

\subsection{Pressure map estimation}

A key capability of the sensor is estimating the contact pressure distribution. The center of pressure and contact shape influence the center of rotation during rotational slip. Additionally, the contact shape and size determine the limit surface and the amount of frictional torque that can be generated for a given normal force \cite{waltersson2024planar}.

\begin{figure}
    \vspace{0.25cm}
    \centering
    \includegraphics[width=0.9\linewidth]{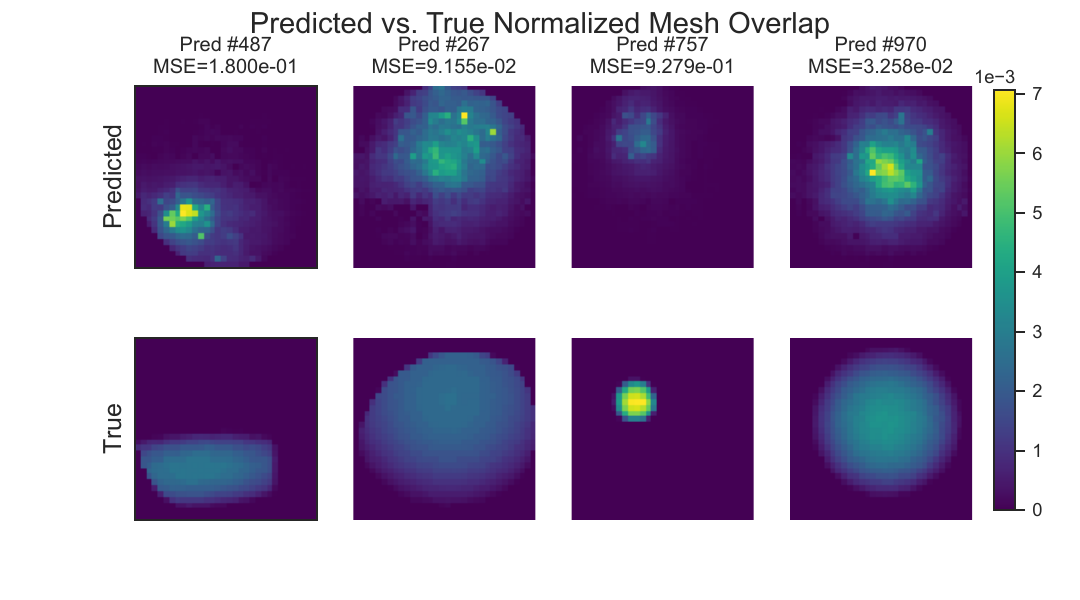}
    \caption{Comparison of predicted and true pressure map from the validation set. }
    \label{fig:preassure_map}
    \vspace{-0.6cm}
\end{figure}

The estimated pressure distribution is visualized in Fig. \ref{fig:preassure_map}, where it can be seen that the sensor is able to reconstruct both the location and size of the \textcolor{blue2}{mesh overlap} region from the Hall-effect sensor readings. The deformable pad contains 12 embedded magnets, which limits the spatial resolution and prevents reconstruction of sharp edges or fine details, unlike high-resolution optical tactile sensors such as GelSight \cite{yuan2017gelsight} or Digit \cite{lambeta2020digit}.  \textcolor{blue2}{However, the proposed tactile sensor operates at 250 Hz and requires less bandwidth than high resolution optical tactile sensing approaches.}

\subsection{Planar velocity sensing}
\begin{figure}[htbp]
    \centering
    \includegraphics[width=0.75\linewidth]{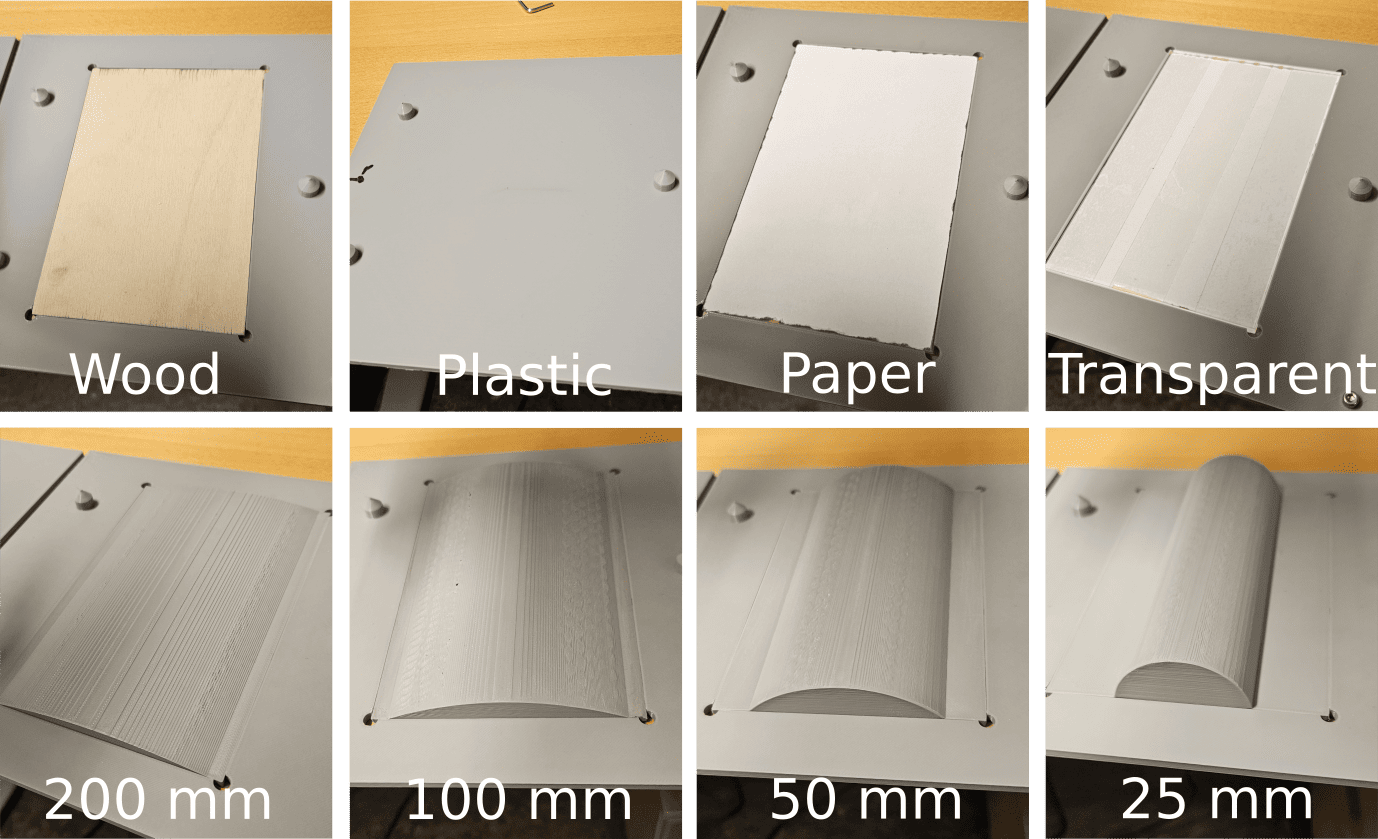}
    \caption{Test surfaces for velocity sensing.}
    \label{fig:surfaces}
    \vspace{-0.3cm}
\end{figure}

\begin{figure}[htbp]
    \centering
    \includegraphics[width=0.7\linewidth]{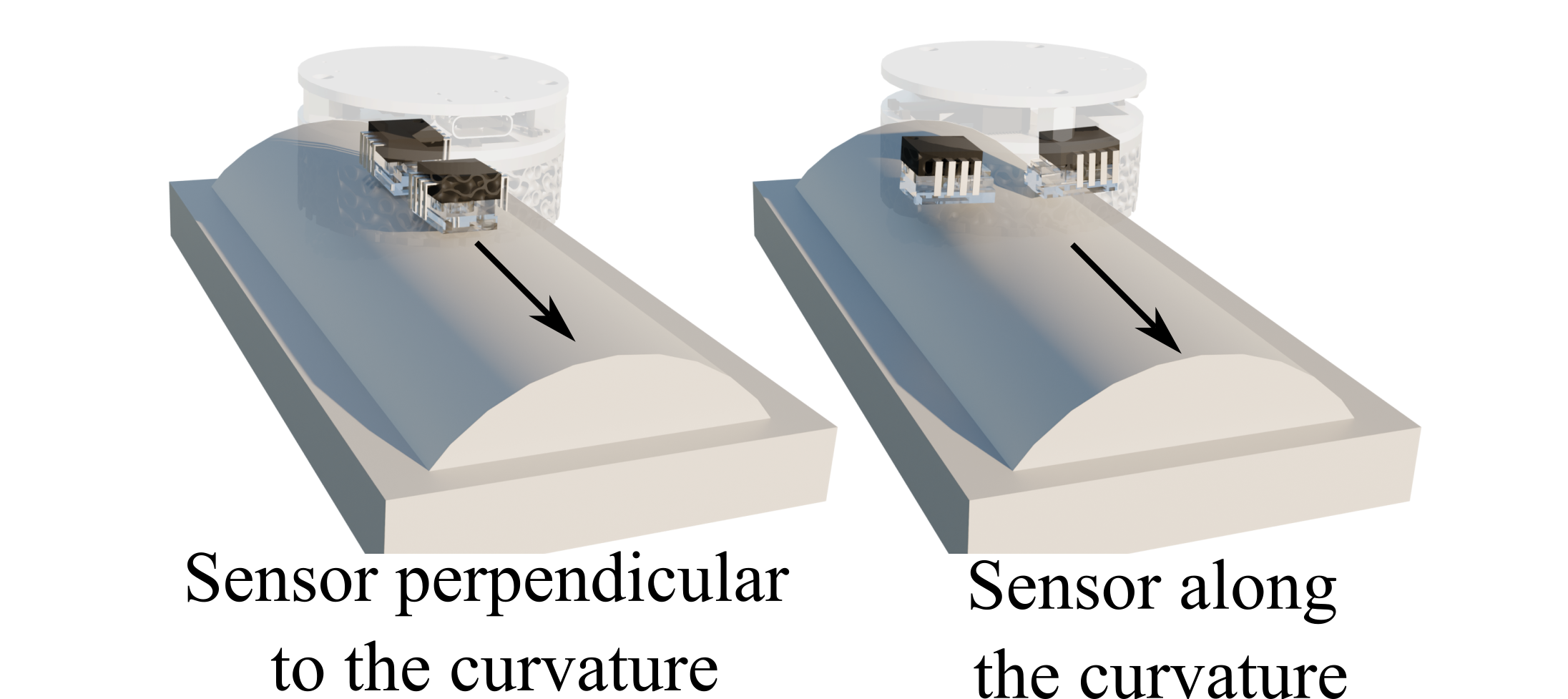}
    \caption{Internal sensor position in relation to object curvature}
    \label{fig:sensor_curvature}
    \vspace{-0.5cm}
\end{figure}

The two sensing modalities described above have previously been integrated into tactile sensors \cite{pattabiraman2025eflesh}. However, \textcolor{blue2}{to the best of our knowledge}, the addition of direct velocity sensing in this form factor has not been realized before. The velocity sensing capabilities and limitations are evaluated across multiple materials and surface curvatures, see Fig. \ref{fig:surfaces}. 
The sensor is tested in three sliding directions: linear motion in the $x$ and $y$ directions, and rotational motion about $\theta$. The motivation for testing both linear directions is illustrated in Fig. \ref{fig:sensor_curvature}.

The normalized relative sliding error is calculated as: \textcolor{blue}{
    $e_i = \frac{\Delta i_\textrm{est}}{\Delta i_\textrm{true}} - 1$
where $\Delta i_\textrm{est}$ is the displacement obtained by integrating the tactile velocity estimate in direction $i \in \{x, y, \theta\}$, and $\Delta i_\textrm{true}$ is the true displacement measured using the forward kinematics of the robot.}
Linear sliding errors are reported in Tab. \ref{tab:tracing_error}. For plastic, wood, and paper surfaces, the errors are small, indicating accurate surface tracking. A bias is observed in the $y$ direction, which could be compensated for by independently calibrating the optical sensors along the $x$ and $y$ axes. The optical sensors have the same strengths and limitations as a computer mouse for tracking different surfaces. Transparent surfaces \textcolor{blue2}{are tracked} poorly, highlighting a limitation of the optical velocity sensor.

Interestingly, for curved surfaces with radii down to 50 mm, the sensor performs well in both linear directions across all tested normal forces, see Tab. \ref{tab:tracing_error}. For highly curved surfaces (25 mm radius), the sensor performs better in one direction than the other. When the optical sensors are aligned along the ridge of the curvature, as illustrated on the left side of Fig. \ref{fig:sensor_curvature}, tracking remains accurate. When the sensors are oriented orthogonally to the ridge, the increased distance between the optical sensor and the surface leads to larger tracking errors. Increasing the normal force slightly improves performance, as the compliant contact pad conforms more closely to the curved surface.

\begin{table*}[t]
\vspace{0.4cm}
\centering
\scriptsize
\begin{tabular}{lcccccccc}
\hline
$f_n$ & Plastic & Wood & Paper & Transparent & Curved 200 & Curved 100 & Curved 50 & Curved 25 \\
\hline
1N $e_x$ & 0.0167 & -0.0076 & -0.0095 & -0.8941 & -0.0092 & 0.0111 & 0.0042 & -0.0154 \\
1N $e_y$ & 0.0489 & 0.0407 & 0.0319 & -0.8643 & -0.0625 & 0.0195 & -0.0387 & -0.9279 \\
\hline
2N $e_x$ & 0.0133 & -0.0060 & 0.0044 & -0.9421 & -0.0172 & 0.0120 & 0.0025 & -0.0087 \\
2N $e_y$ & 0.0601 & 0.0409 & 0.0303 & -0.9574 & -0.0408 & 0.0193 & -0.0241 & -0.8622 \\
\hline
4N $e_x$ & 0.0146 & -0.0081 & 0.0019 & -0.9805 & -0.0235 & 0.0159 & 0.0056 & -0.0020 \\
4N $e_y$ & 0.0652 & 0.0496 & 0.0504 & -0.9931 & -0.0330 & 0.0215 & -0.0176 & -0.7353 \\
\hline
8N $e_x$ & 0.0182 & -0.0062 & -0.0048 & -0.9961 & -0.0312 & 0.0106 & 0.0035 & -0.0020 \\
8N $e_y$ & 0.0691 & 0.0468 & 0.0376 & -0.9983 & -0.0275 & 0.0218 & -0.0121 & -0.5672 \\
\hline
\end{tabular}
\caption{Normalized relative linear tracking error for different materials (Fig. \ref{fig:surfaces}) and normal forces}
\label{tab:tracing_error}
\vspace{-0.5cm}
\end{table*}

\begin{table*}[b]
\centering
\scriptsize
\begin{tabular}{lcccccccc}
\hline
 $f_n$ & Plastic & Wood & Paper & Transparent & Curved 200 & Curved 100 & Curved 50 & Curved 25 \\
\hline
1N $e_\psi$ & 0.0113 & 0.0263 & 0.0196 & -0.9029 & -0.0485 & 0.0031 & -0.0194 & -0.2428 \\
\hline
2N $e_\psi$ & 0.0231 & 0.0355 & 0.0217 & -0.9389 & -0.0131 & 0.0144 & -0.0094 & -0.2086 \\
\hline
4N $e_\psi$ & 0.0367 & 0.0391 & 0.0287 & -0.9808 & -0.0117 & 0.0180 & 0.0049 & -0.1425 \\
\hline
8N $e_\psi$ & 0.0289 & 0.0332 & 0.0316 & -0.9999 & -0.0137 & 0.0212 & 0.0134 & -0.1195 \\
\hline
\end{tabular}
\caption{Normalized relative yaw tracking error for different materials (Fig. \ref{fig:surfaces}) and normal forces}
\label{tab:tracing_error_yaw}
\end{table*}

Similar trends are observed for rotational tracking, see Tab. \ref{tab:tracing_error_yaw}. For flat surfaces and curvatures down to 50 mm radius, accurate rotational tracking is achieved on materials that also perform well for linear sliding. For highly curved surfaces, tracking accuracy degrades, but improves modestly with increased normal force. As with linear motion, transparent surfaces are not reliably tracked.

\subsection{In-hand sliding}

\begin{figure}[tbp]
    \centering
    \includegraphics[width=0.65\linewidth]{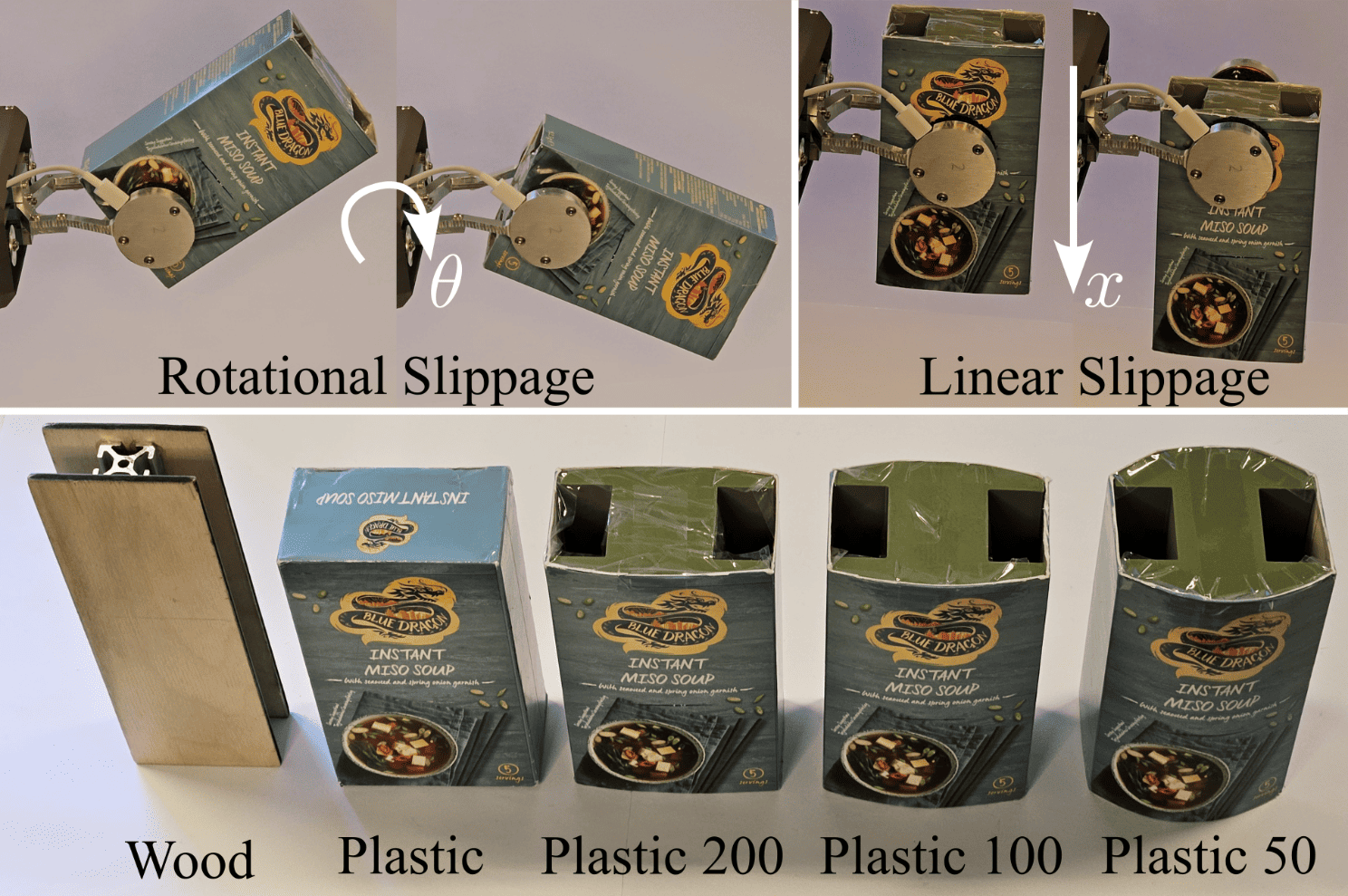}
    \caption{In-hand sliding experiments (rotational and linear) on objects with varying curvature and materials; object specifications are given in \cite{waltersson2024inhand}. 
    }
    \label{fig:slip_figs}
\end{figure}

To evaluate sensor performance during in-hand sliding manipulation, the sensors are mounted on a custom gripper (Fig. \ref{fig:overview}), which is a repackaged version of the gripper presented in \cite{waltersson2024inhand}. Two types of in-hand slip experiments are conducted: rotational and linear trajectory tracking (Fig. \ref{fig:slip_figs}). The \textcolor{blue}{closed-loop} slip-aware controllers are adopted from \cite{waltersson2024inhand}, with parameters manually retuned for the new system.  This work focuses on sensor performance rather than controller design. \textcolor{blue}{During slip, the primary sensing modality for feedback is velocity, whereas during stiction, it is force.} Throughout all experiments, only \textcolor{blue}{measurements from} one of the two sensors mounted on the gripper is used for slip-aware control.

\begin{figure}[tbp]
    \centering
    \includegraphics[width=0.8\linewidth]{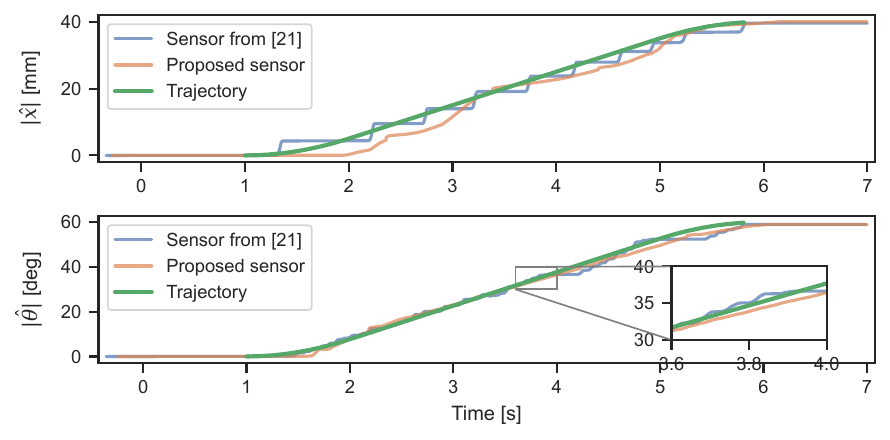}
    \caption{Trajectory-following experiments from \cite{waltersson2024inhand} using the proposed sensors on a custom gripper with a plastic object; see Fig. \ref{fig:slip_figs} for the corresponding sliding primitives. \textcolor{blue}{The trajectory-following results are measured using the integrated velocity sensors.}}
    \label{fig:trajectory_following}
    \vspace{-0.6cm}
\end{figure}

The sliding behavior of the proposed sensors, which feature a deformable contact pad, is compared to that of the sensors in \cite{waltersson2024inhand}, which use a rigid contact pad. As shown in Fig. \ref{fig:trajectory_following}, the deformable pad minimizes stick–slip effects, resulting in smoother motion.
This effect is more pronounced in linear sliding but also present in rotation.

Each object shown in Fig. \ref{fig:slip_figs} is tested ten times for both rotational and linear sliding, resulting in a total of 100 experiments. No instances of object drop occurred. The \textcolor{blue}{endpoint} ground-truth rotation angle is measured using a digital inclinometer, while the linear displacement is measured using a caliper. These measurements are compared to the displacement estimated by integrating the velocity sensor output. The results are benchmarked against those reported in~\cite{waltersson2024inhand} and summarized in Table~\ref{tab:slip_test}.

For rotational sliding, the proposed sensors significantly outperform those in \cite{waltersson2024inhand} for the plastic object. The rigid contact pads used in \cite{waltersson2024inhand} did not provide sufficient torque for the wooden object and were not evaluated on curved surfaces. In contrast, the proposed sensors successfully perform rotational sliding on all objects shown in Fig.~\ref{fig:slip_figs}.

For linear sliding, the sensors in~\cite{waltersson2024inhand} achieve slightly better performance on flat objects; however, those results rely on outlier rejection and sensor fusion  across both sensors. Notably, for curved objects, the rigid sensors fail to maintain the appropriate sensing distance due to their inability to conform to the surface geometry. The proposed sensors, with their deformable contact pad and improved sensor placement, effectively adapt to curved surfaces and enable reliable linear in-hand sliding for all tested objects.

\begin{table}[]
    \vspace{0.3cm}
    \centering
    \resizebox{\columnwidth}{!}{%
    \begin{tabular}{c|c|c|c|c}
    \hline
        Object & $\theta - \hat{\theta}$ & $\theta - \hat{\theta}$ \cite{waltersson2024inhand} & $x - \hat{x}$ & $x - \hat{x}$ \cite{waltersson2024inhand} \\
         &  ($\mu$, $\sigma$) & ($\mu$, $\sigma$)  &  ($\mu$, $\sigma$) & ($\mu$, $\sigma$)\\
         \hline
         Plastic flat  ($60^\circ$, 40 mm)& (2.46, 2.77) & (17.1, 7.0) & (4.69, 1.58) & (1.84, 0.73) \\ \hline
         Wood flat ($60^\circ$, 40 mm) & (-7.08, 3.54) & NA &  (2.36, 2.09) & (0.70, 0.57)  \\ \hline
         Plastic 200 ($60^\circ$, 20 mm) & (0.44, 4.25) & NT & (-0\textcolor{blue2}{.}84, 2.10) & (36.59, 7.74) \\ \hline
         Plastic 100 ($60^\circ$, 20 mm) & (-4.21, 8.16) & NT & (2.08, 4.79) & NA \\ \hline
         Plastic 50 ($60^\circ$, 20 mm) & (-2.32, 4.23) & NT & (1.57, 1.50) & NA \\ \hline
    \end{tabular}}
    \caption{Position tracking error in linear and rotational directions, compared with the sensors presented in \cite{waltersson2024inhand}. All trajectories have a duration of 5 s. NA denotes unavailable due to system limitations; NT denotes not tested.}
    \label{tab:slip_test}
    \vspace{-0.6cm}
\end{table}

These results demonstrate that the proposed tactile sensor is capable of accurately measuring sliding displacement while simultaneously estimating 6-DoF F/T and contact pressure distributions, all within a compact form factor and easily manufacturable tactile sensor. The sensor significantly outperforms the sensors in \cite{waltersson2024inhand} on curved surfaces and provides better slip-stick dynamics due to the deformable contact pad.

\section{Conclusions}

In this paper, we introduce a tactile sensor that combines sensing modalities not previously integrated into a single device, targeting in-hand sliding manipulation. Experiments demonstrate its ability to estimate 6-DoF F/T, \textcolor{blue2}{approximate} contact pressure distributions, and sliding velocity. The velocity tracking exhibits similar strengths and limitations to computer mouse sensors, performing well on most \textcolor{blue}{diffuse material} surfaces. The main contribution lies in the integration of sensing modalities with a deformable contact pad, rather than in the performance of individual modalities. 

The sensor is evaluated on simple in-hand manipulation tasks using a fast parallel gripper. Future work will focus on leveraging these modalities for in-hand perception, \textcolor{blue}{contact estimation}, more advanced sliding manipulation with planning, \textcolor{blue}{alternative contact pads, including approaches such as \cite{anyskin}, viscoelastic compensation}, multimodal sensor fusion \textcolor{blue}{and fallback strategies}, and the combined control of the robotic arm and in-hand slip-aware control.

\addtolength{\textheight}{-0cm}   




\bibliographystyle{IEEEtran}
\bibliography{IEEEabrv,references}

\end{document}